\documentclass[conference]{IEEEtran}

\usepackage{multicol}
\usepackage{amsmath,amssymb,cite,multirow,subfigure}
\usepackage{graphicx}
\usepackage{url}
\usepackage[linesnumbered,ruled, vlined]{algorithm2e}
\usepackage{algorithmic}
\usepackage{makecell}

\usepackage{booktabs}
\usepackage{threeparttable}

\begin{document}
%
\title{Intra-Layer Nonuniform Quantization of Convolutional Neural Network}

\author{\IEEEauthorblockN{Fangxuan Sun, Jun Lin, and Zhongfeng Wang}
\IEEEauthorblockA{School of Electronic Science and Engineering, Nanjing University, China\\
Email: njusunfx@gmail.com, \{jlin, zfwang\}@nju.edu.cn}}

\maketitle

\begin{abstract}

Deep convolutional neural network (DCNN) has
achieved remarkable performance on object detection and speech
recognition in recent years. However, the excellent performance of
a DCNN incurs high computational complexity and large memory
requirement. In this paper, an equal distance nonuniform quantization
(ENQ) scheme and a K-means clustering nonuniform
quantization (KNQ) scheme are proposed to reduce the required memory
storage when low complexity hardware or software implementations are
considered. For the VGG-16 and the AlexNet, the proposed
nonuniform quantization schemes reduce the number of
required memory storage
by approximately 50\% while achieving almost
the same or even better classification accuracy compared to the state-of-the-art
quantization method. Compared to the ENQ scheme, the
proposed KNQ scheme provides a better tradeoff when higher
accuracy is required.

\end{abstract}


\section{Introduction}
\label{sec:intro}

Convolutional neural network (CNN) has been widely used in recent years due to its remarkable performance in object detection and speech recognition. It was shown in~\cite{krizhevsky2012imagenet,simonyan2014very,he2015deep} that deep convolutional neural network (DCNN) achieves remarkable accuracy in image classification. Recently, lots of research efforts~\cite{srivastava2014dropout:,ioffe2015batch} have been devoted to improve the performance of DCNNs and reduce the gradient vanishing~\cite{erhan2009the, glorot2010understanding} in the training process. Due to the fantastic performance of CNN, many computer vision applications also employ CNN to improve their performance. For example, CNN has achieved great performance in image annotation~\cite{karpathy2015deep}, visual QA system~\cite{antol2015vqa}, 3D interpreter~\cite{wu2016single} and many other areas. Moreover, CNN was applied in speech recognition in~\cite{amodei2015deep} and was shown to achieve higher accuracy compared to previous methods.

DCNN performs well at the cost of dramatically increased computational complexity. Hence, efficient hardware implementation of these networks for real time processing is very challenging. The deep structure of DCNNs not only increases the computational complexity, but also incurs significant storage requirement. The weights and activations dominate the overall storage. Activations are the pixels of feature maps in a CNN. For the VGG-16~\cite{simonyan2014very} net, the memory required to store all weights and activations is around 2Gb and 200Mb, respectively, when each weight and activation are half-precision floating-point numbers.
In~\cite{han2015deep}, the number of trained parameters of DCNNs has been reduced to less than $5\%$ of their original size. 

One important approach to reduce the storage requirement is to replace the full or half-precision floating-point number with fixed-point number for the computations of DCNNs. Researchers have proposed various fixed-point quantization schemes for the activations of DCNNs. A cross layer nonuniform quantization (CLNQ) scheme was proposed in~\cite{lin2015fixed} to minimize the number of bits required to store all activations. In~\cite{moons2016energy}, a nonuniform quantization scheme and the approximate computing technique were combined together to reduce the power consumption of a CNN. However, the number of required memory bits to store all activations is still very large even with these fixed-point quantization schemes.

In this paper, we focus on the efficient quantization of activations in a DCNN. The main contributions of this work are as follows.
\begin{itemize}
\item Two intra-layer nonuniform quantization (ILNQ) schemes, equal distance intra-layer nonuniform quantization (ENQ) scheme and K-means clustering based intra-layer nonuniform quantization (KNQ), are proposed for the quantization of activations in DCNNs. Compared to the state-of-the-art quantization scheme in~\cite{lin2015fixed}, the proposed ILNQ schemes reduce the number of required memory bits to store all activations while maintaining almost the same accuracy.
\item Compared to the ENQ scheme, the KNQ scheme improves the accuracy and slightly reduces the number of required memory bits at the cost of small hardware overhead when hardware implementations are considered. The KNQ scheme provides a tradeoff between hardware complexity and accuracy.
\item Both the ENQ and KNQ schemes are applied to the quantization of VGG-16 and AlexNet~\cite{krizhevsky2012imagenet} with the ILSVRC-2012~\cite{ILSVRC15} data set. Compared to the quantization scheme in~\cite{lin2015fixed}, it is demonstrated that both of the ENQ and KNQ schemes reduce the number of required memory bits to store all activations by approximately 50\% while achieving almost the same or even better accuracy. Compared to the floating-point implementation, the accuracy loss is less than 2\% for both of the ENQ and KNQ schemes.
\end{itemize}

The rest of the paper is organized as follows. Related works are discussed in Section~\ref{sec: related}. The proposed intra-layer nonuniform quantization schemes are presented in Section~\ref{sec: quan}. The comparisons of different quantization methods and related discussions are provided in Section~\ref{sec: comp}. At last, the conclusions are drawn in Section~\ref{sec: conclusion}.

\section{Related Work}\label{sec: related}
The quantization of DCNNs has been widely discussed in the open literatures. These works can be categorized into weight compression and activation quantization.
Many previous works focus on weight compression of convolutional and fully-connect layers. A partial pruning algorithm was proposed in~\cite{anwar2015structured} to reduce the memory required by all weights. A compression method, which consists of pruning, quantization and Huffman coding, was proposed in~\cite{han2015deep} to compress the weights of a DCNN. The coding technology~\cite{gong2014compressing} was also employed to perform the quantization of DCNN as well. In~\cite{denton2014exploiting}, it was showed that the weight of fully-connected layers can be compressed by truncated singular value decomposition.  A hash trick was proposed in~\cite{chen2015compressing} to compress the DCNN. Moreover, binary weights were employed in~\cite{courbariaux2015binaryconnect:,rastegari2016xnor-net:} to reduce both computational complexity and storage requirement at the cost of certain accuracy loss. On the other hand, the quantization of activations has rarely been discussed. In~\cite{lin2015fixed}, the signal-to-quantization noise ratio (SQNR) metric was employed to compute the number of quantization bits for each activation. The quantization of activations in the fully-connect layer was discussed in~\cite{hwang2014fixed-point}.


\section{Proposed Intra-Layer Nonuniform Quantization of DCNNs}
\label{sec: quan}

\begin{figure} [hbt]
\centering
  \includegraphics[width=3in]{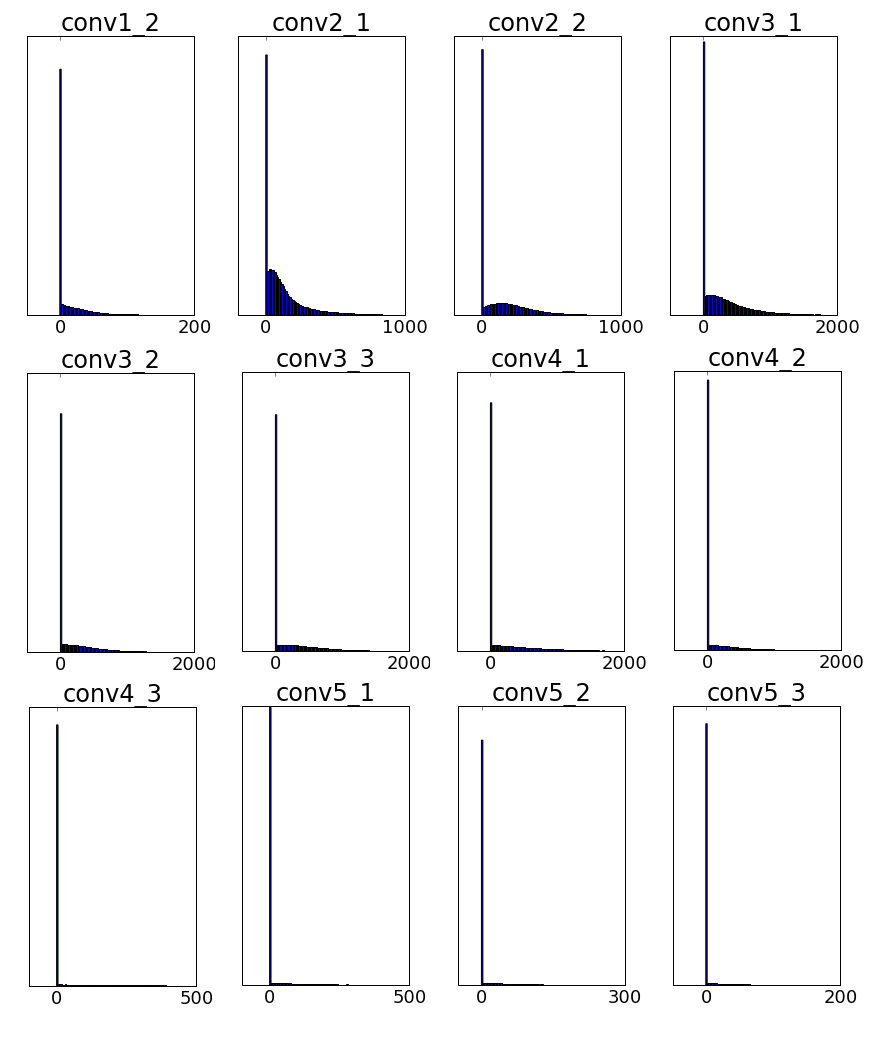}
  \caption{Distribution of activations in VGG-16.}\label{fig: vgg-16}
\end{figure}

The proposed ILNQ of DCNN is based on the empirical distributions of activations of each convolutional layer. The distribution of activation data for each layer of a CNN on the CIFAR-10 benchmark was shown in~\cite{lin2015fixed}. According to their experiments, the distribution of both activations and weights in most layers are roughly Gaussian distribution. To find out whether the distribution of activations in a deeper CNN is Gaussian-like distribution as well, we check the data of some deeper networks such as VGG-16 and AlexNet on the data set of ILSVRC-2012. Take the VGG-16 as an example, the distribution of activations with each layer is shown in Fig.~\ref{fig: vgg-16}, where the horizontal and vertical coordinates denote the magnitude and the corresponding probability, respectively. Note that the activations after ReLU are quantized in this paper. Since the ReLU function has non-negative output, all activations discussed in this paper are non-negative. As shown in Fig.~\ref{fig: vgg-16}, for VGG-16, the distribution of activations in each convolutional layer is Gaussian-like. Besides, most of the activations are zero for each convolutional layer shown in Fig.~\ref{fig: vgg-16}. Similar observations are obtained for the AlexNet based on the same data set.

\subsection{Proposed ENQ and KNQ Schemes}
The quantization process can be viewed as a mapping of activations to a set of quantization points (QPs), where each QP corresponds to a fixed-point value that will be used in the future computations. Let $\{P_0,P_1,\cdots,P_{N-1}\}$ denote a set of QPs with $N$ elements. For $i=0,1,\cdots,N-1$, let $V_i$ denote the corresponding fixed-point value associated with $P_i$. The quantization of an activation $a$ is the following mapping:
\begin{equation}\label{equ: mapping}
a \rightarrow P_k,
\end{equation}
where $k\in\{0,1,\cdots,N-1\}$. The storage of an quantized activation requires $K=\lceil\log_2N\rceil$ bits. Let $K'$ denote the number of bits used to store a $V_i$ for $i=0,1,\cdots,N-1$. It is possible that $K \neq K'$.

As shown in Fig.~\ref{fig: vgg-16}, it can be concluded that most activations within a layer are relatively small. It is reasonable to assign more QPs to represent the smaller activations without incurring obvious degradation of accuracy. In more detail, the proposed two ILNQ schemes are described as follows.
\begin{itemize}
\item For uniform quantization, i.e., each activation in a DCNN is quantized with $q$ bits ($2^q$ QPs in total). Note that all activations are non-negative since we are storing activations after the ReLU step. Let $F$ denote the number of fractional bits in the $q$-bit uniform quantization. For the uniform quantization mentioned below in this paper, $V_i=i2^{-F}$ for $i=0,1,\cdots,2^q-1$. For a DCNN, let $A$ and $A'$ denote the accuracy with floating and $q$-bit uniform quantization, respectively. The minimal value of $q$ (denoted as $q'$) is calculated such that $A-A'\leqslant \delta$, where $\delta$ is a small positive number determined by the corresponding data set and application.
\item For activations within each layer $i$, the proposed ENQ scheme employs an $E_i$-bit nonuniform quantization scheme, where $E_i\leqslant q_m$. For the proposed $E_i$-bit ENQ scheme, there are $2^{E_i}$ QPs, $P_{i,0}$, $P_{i,1},\cdots,P_{i,2^{E_i}-1}$, where $P_{i,k}$ corresponds to the fixed-point value $V_{i,k}=k2^{q_m-E_i}2^{-F}$. Here, $k=0,1,\cdots,2^{E_i}-1$. Supposing the magnitude of an activation is $x$, the ENQ scheme quantizes it to $\lfloor \frac{x}{2^{q_m-E_i}}\rfloor$ if $x\leqslant2^{q_m}-1$. Otherwise, $x$ is quantized to $2^{E_i}-1$. Note that each activation within layer $i$ is stored using $E_i$ bits. When an activation is needed to participate in the convolutional computations in the next layer, it should be converted to a $q_m$-bit activation first based on the relationship between QPs and fixed-point values.
    In this paper, exhaustive search is employed to find the minimal $E_i$ for each layer $i$ such that the resulting accuracy is close to $A'$.
\item Considering that the distribution of activations within a convolutional layer is Gaussian-like, the proposed ENQ scheme does not take full advantage of the distribution. The proposed KNQ scheme employs K-means clustering method to assign each QP with a fixed-point value. 
    Let $S$ denote the set of all activations within convolutional layer $i$. Suppose $T_i$ bits are used to quantize all these activations. Let $S_0,S_1,\cdots,S_{2^{T_i}-1}$ be $2^{T_i}$ non-overlap sets which divide the whole set $S$, where $\cup_{k=0}^{2^{T_i}-1}S_k=S$. For a given $T_i$, the proposed KNQ scheme first finds $2^{T_i}$ fixed-point values, $d_0, d_1,\cdots,d_{2^{T_i}-1}$, by solving the following problem with the K-means clustering method:
    \begin{equation}\label{equ: mld}
    \min_{d_0, d_1,\cdots,d_{2^{T_i}-1}}\sum_{k=0}^{2^{T_i}-1}\sum_{j=0}^{|S_k|}|s_{k,j}-d_k|^2,
    \end{equation}
    where $|S_j|$ denotes the number of elements in $S_j$ and $s_{k,j}\in S_k$. Once these $2^{T_i}$ values are determined, $x$ is quantized to $k$ if $d_k\leqslant x <d_{k+1}$. If $x\geqslant d_{2^{T_i}-1}$, $x$ is quantized to $2^{T_i}-1$. For $k=0,1,\cdots,2^{T_i}-1$, quantization point $P_{i,k}$ corresponds to $d_k$. Similar to the ENQ method, a quantized activation will be mapped to the corresponding fixed-point value first when it is needed in the future computation.
\end{itemize}
For the KNQ scheme, in order to reduce the accuracy loss, the set $S$ is pre-processed in this paper. All activations larger than a threshold value $M$ is saturated to $M$. In this paper, $M=(2^{q_m}-1)2^{-F}$, since the uniform quantization demonstrates that considering larger $q_m$ is not necessary.

\subsection{Proposed Data Conversion Units}
\begin{figure} [hbt]
\centering
  \includegraphics[width=2.9in]{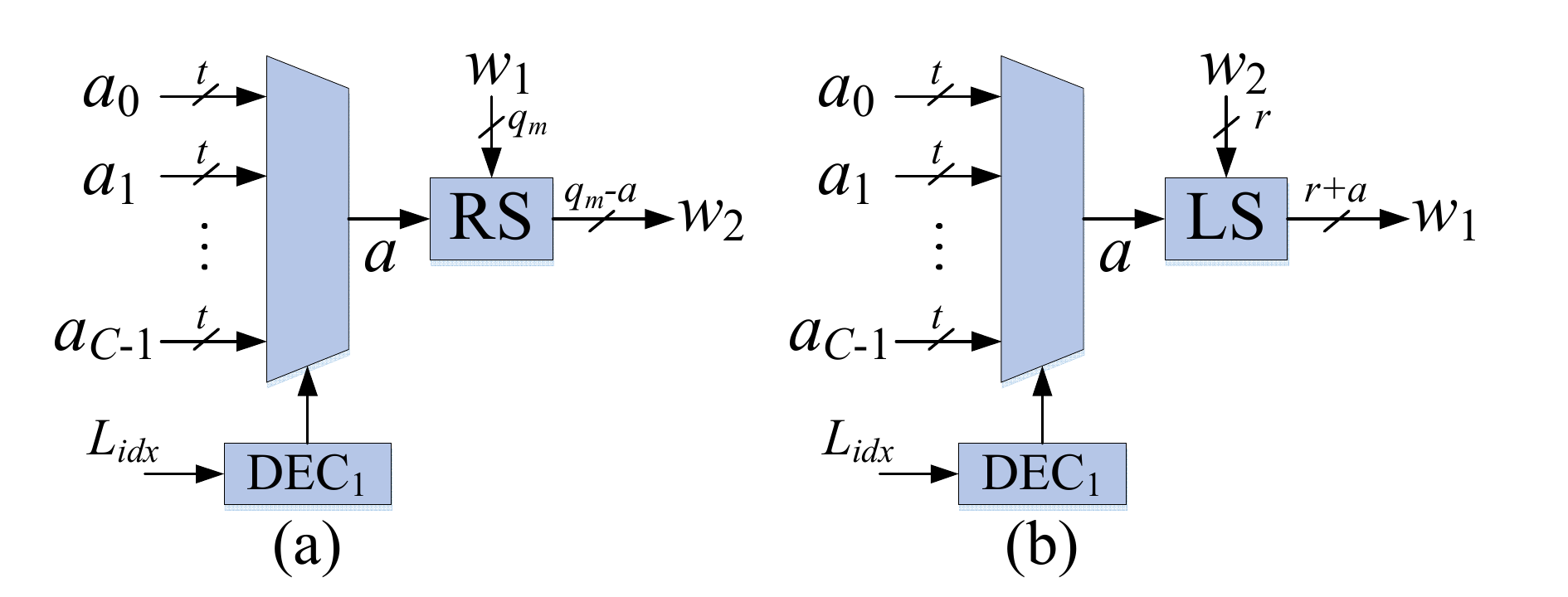}
  \caption{QU and conversion unit for ENQ.}\label{fig: dce}
\end{figure}

\begin{figure} [hbt]
\centering
  \includegraphics[width=2.6in]{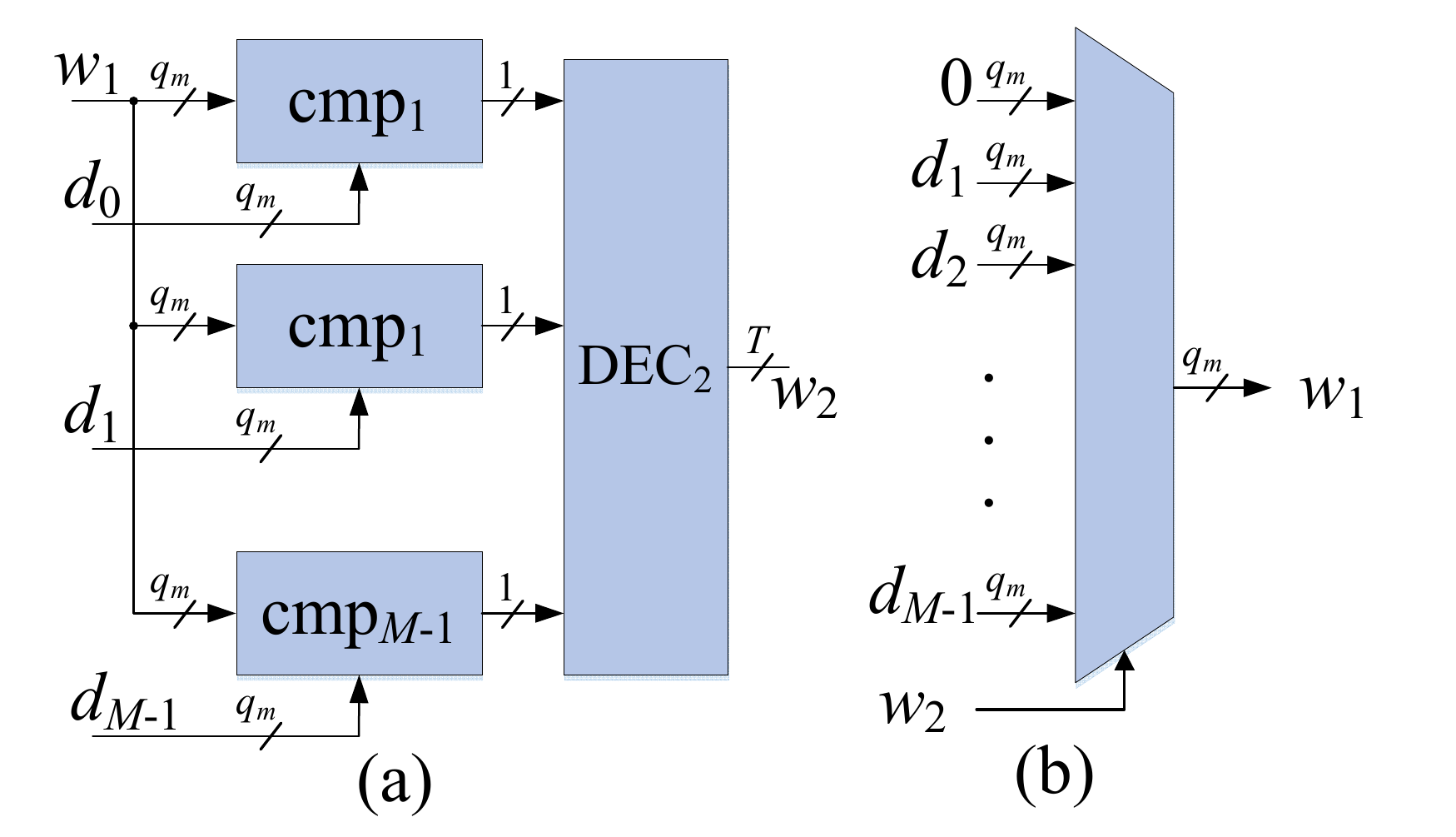}
  \caption{QU and conversion unit for KNQ.}\label{fig: dck}
\end{figure}

In terms of hardware implementation, compared to the uniform quantization scheme and the quantization scheme in~\cite{lin2015fixed}, the proposed ENQ and KNQ schemes need extra circuits to convert between QPs and corresponding fixed-point values. In this paper, the efficient data conversion hardware architectures for the ENQ and KNQ schemes are proposed in Figs.~\ref{fig: dce} and~\ref{fig: dck}, respectively. Let $w_1$ and $w_2$ denote a fixed-point value and the corresponding quantization point. For the ENQ scheme, the quantization unit of ENQ (QE) is in Fig.~\ref{fig: dce}(a), where $C$ is the number of different quantization bits and $L_idx$ is the index of the convolutional layer. The DEC1 unit takes $L_idx$ as input to generate selection signal for the multiplexor. The right shift (RS) unit shift the input to right by $a$ bits. The conversion unit for ENQ (CE) is shown in Fig.~\ref{fig: dce}(b), where the left shift (LS) unit shifts the input to left by $a$ bits.

Let $M$ denote the number of centroid values generated by the K-means clustering method. The QC of KNQ (QK) is shown in Fig.~\ref{fig: dck}(a), where cmp is a comparator. DEC$_2$ is a priority encoder unit which generates the quantized value. As shown in Fig.~\ref{fig: dck}(b), the conversion unit for KNQ (CK) can be implemented with a multiplexor which selects out the corresponding centroid values for the input QP.
 


%
%
%

\section{Numerical Results and Comparisons}
\label{sec: expe}

In this paper, based on the ILSVRC-2012 data set, the proposed ILNQ schemes are applied on the VGG-16 and AlexNet. The VGG-16 achieves remarkable performance on image classification and its deep structure can be modified to adapt to various applications. Moreover, the size of activations in the VGG-16 is very huge. Hence, VGG-16 is a typical representation of DCNNs. For all the implementations in this paper, all the weights and bias take a 14-bit uniform quantization.

\subsection{Uniform Quantization and Cross-Layer Nonuniform Quantization}\label{fixed::uniform}
We first perform the uniform quantization of VGG-16 according to the distribution shown in Fig.~\ref{fig: vgg-16}. As shown in Table~\ref{tab: uniform}, the accuracy of the full precision is $88.5\%$. We use the same bit-width for all layers without fine-tuning. According to our observations, some activations are even larger than $15000$. However, 12 bits are enough for the uniform quantization of all activations.

\begin{table}[hbt]
  \centering
  \caption{Uniform Quantization of VGG-16}
  \label{tab: uniform}
  \footnotesize
  \begin{tabular}{c|c|c|c|c|c|c}
    \hline
   Quantization bit  & 14 & 13 & 12 & 11 & 10 & 9  \\ \hline\hline
   Top5 Accuracy($\%$)      &    $88.4$   &   $88.4$   &   $88.5$    &    $87.0$    &    $79.6$   &   $56.0$   \\\hline
  \end{tabular}
\end{table}

\begin{table*}[hbt]
  \centering
  \caption{ENQ of VGG-16 Part 1}
  \label{tab:discrete1}
  \begin{threeparttable}
  \footnotesize
  \begin{tabular}{c||c||c|c|c|c|c|c|c|c|c|c}
    \hline
    Top5 Accuracy($\%$) & Index &conv1$\_$1 & conv1$\_$2 &  conv2$\_$1 & conv2$\_$2 & conv3$\_$1 & conv3$\_$2 & conv3$\_$3 & conv4$\_$1 & conv4$\_$2 & conv4$\_$3   \\ \hline\hline
     $56.60$ & 1 &8   &4    &5   & 5&5&5 & 5&5&5 & 5 \\ \hline
     $79.43$ & 2 &8   &4    &5   & 5&5&5 & 5&5&5 & 5 \\ \hline
     $83.46$ & 3 &8   &4    &5   & 5&5&5 & 5&5&5 & 5 \\ \hline
     $85.81$ & 4 &8   &5    &6   & 4&6&6 & 6&6&6 & 6 \\ \hline
     $86.19$ & 5 &8   &5    &6   & 5&5&4 & 6&6&6 & 6 \\ \hline
     $86.25$ & 6 &8   &5    &6   & 5&5&5 & 5&5&5 & 5 \\ \hline
  \end{tabular}
    \begin{tablenotes}[para,flushleft]
  \end{tablenotes}
  \end{threeparttable}
\end{table*}

\begin{table*}[hbt]
  \centering
  \caption{ENQ of VGG-16 Part 2}
  \label{tab:discrete2}
  \begin{threeparttable}
  \footnotesize
  \begin{tabular}{c||c||c|c|c|c|c|c}
    \hline
    Top5 Accuracy($\%$) & Index &conv5$\_$1 & conv5$\_$2 &  conv5$\_$3 & fc6 & fc7 & fc8    \\ \hline\hline
     $56.60$ & 1 &5   &4    &4   & 3&2&2 \\ \hline
     $79.43$ & 2 &5   &4    &4   & 4&2&2 \\ \hline
     $83.46$ & 3 &5   &4    &4   & 6&3&2\\ \hline
     $85.81$ & 4 &6   &5    &5   & 6&3&2\\ \hline
     $83.46$ & 5 &6   &5    &5   & 6&3&2\\ \hline
     $86.25$ & 6 &6   &5    &5   & 6&3&2\\ \hline
  \end{tabular}
    \begin{tablenotes}[para,flushleft]
  \end{tablenotes}
  \end{threeparttable}
\end{table*}

\begin{table*}[hbt]
  \centering
  \caption{K-means}
  \label{tab:centroid}
  \begin{threeparttable}
  \footnotesize
  \begin{tabular}{c|c|c|c|c|c|c|c|c|c|c|c|c|c|c|c|c}
    \hline
    & $d_{0}$ & $d_{1}$ & $d_{2}$ & $d_{3}$ & $d_{4}$ & $d_{5}$ & $d_{6}$ & $d_{7}$ & $d_{8}$ & $d_{9}$ & $d_{10}$ & $d_{11}$ & $d_{12}$ & $d_{13}$ & $d_{14}$ & $d_{15}$  \\ \hline\hline

    conv$1\_2$ & 0 & 0 & 0 & 0 & 0 & 0 & 0 & 0 & 0 & 0 & 0 & 0 & 0  & 1 & 3 & 6  \\ \hline
    conv$2\_1$ & 0 & 11 & 22 & 35 & 49 & 64 & 80 & 97 & 115 & 134 & 154 & 176 & 200  & 226 & 254 & 284  \\ \hline
    conv$2\_2$ & 0 & 23 & 46 & 69 & 92 & 115 & 138 & 161 & 185 & 210 & 236 & 263 & 292 & 323  & 356 & 391  \\ \hline
    conv$3\_1$ & 0 & 45 & 91 & 138 & 186 & 235 & 285 & 338 & 394 & 453 & 516 & 583 & 653  & 726 & 803 & 886  \\ \hline
    conv$3\_2$ & 0 & 45 & 90 & 136 & 183 & 232 & 282 & 333 & 386 & 441 & 498 & 558 & 621  & 689 & 763 & 843  \\ \hline
    conv$3\_3$ & 0 & 48 & 97 & 146 & 196 & 247 & 299 & 352 & 407 & 464 & 523 & 585 & 649  & 716 & 786 & 860  \\ \hline
    conv$4\_1$ & 0 & 55 & 113 & 172 & 233 & 297 & 361 & 426 & 494 & 566 & 641 & 720 & 801  & 886 & 973 & 1066  \\ \hline
    conv$4\_2$ & 0 & 44 & 88 & 133 & 179 & 227 & 277 & 329 & 385 & 443 & 504 & 567 & 632  & 703 & 779 & 862  \\ \hline
    conv$4\_3$ & 0 & 35 & 71 & 109 & 149 & 191 & 235 & 282 & 333 & 388 & 447 & 513 & 586  & 667 & 760 & 867  \\ \hline
    conv$5\_1$ & 0 & 28 & 58 & 91 & 129 & 169 & 215 & 267 & 323 & 384 & 452 & 527 & 612  & 713 & 824 & 952  \\ \hline
    conv$5\_2$ & 0 & 16 & 34 & 54 & 77 & 103 & 132 & 165 & 203 & 247 & 296 & 356 & 425  & 499 & 585 & 687  \\ \hline
    conv$5\_3$ & 0 & 10 & 22 & 36 & 52 & 71 & 93 & 119 & 148 & 177 & 211 & 248 & 288  & 330 & 370 & 415  \\ \hline

     & $d_{16}$ & $d_{17}$ & $d_{18}$ & $d_{19}$ & $d_{20}$ & $d_{21}$ & $d_{22}$ & $d_{23}$ & $d_{24}$ & $d_{25}$ & $d_{26}$ & $d_{27}$ & $d_{28}$ & $d_{29}$ & $d_{30}$ & $d_{31}$  \\ \hline\hline

    conv$1\_2$ & 10 & 15 & 21 &28 & 36 & 45 & 56 & 69 & 85 & 105 & 128 & 156 & 191  & 234 & 290 & 373  \\ \hline
    conv$2\_1$ & 316 & 350 & 387 & 426 & 468 & 514 & 563 & 616 & 674 & 739 & 814 & 901 & 1015  & 1162 & 1407 & 1815  \\ \hline
    conv$2\_2$ & 429 & 470 & 517 & 569 & 628 & 695 & 772 & 863 & 970 & 1094 & 1245 & 1432 & 1666 & 1977  & 2425 & 3082   \\ \hline
    conv$3\_1$ & 977 & 1075 & 1181 & 1299 & 1428 & 1570 & 1730 & 1908 & 2092 & 2286 & 2517 & 2760 & 3041  & 3394 & 3834 & 4096  \\ \hline
    conv$3\_2$ & 930 & 1024 & 1128 & 1243 & 1368 & 1505 & 1655 & 1818 & 2000 & 2216 & 2466 & 2730 & 3013  & 3368 & 3821 & 4096  \\ \hline
    conv$3\_3$ & 940 & 1027 & 1120 & 1222 & 1334 & 1457 & 1598 & 1757 & 1941 & 2151 & 2391 & 2662 & 2983  & 3357 & 3820 & 4096  \\ \hline
    conv$4\_1$ & 1164 & 1270 & 1384 & 1505 & 1635 & 1775 & 1927 & 2094 & 2266 & 2473 & 2703 & 2943 & 3215  & 3525 & 3893 & 4096  \\ \hline
    conv$4\_2$ & 956 & 1062 & 1178 & 1302 & 1437 & 1584 & 1752 & 1946 & 2152 & 2397 & 2653 & 2959 & 3250  & 3551 & 3888 & 4096  \\ \hline
    conv$4\_3$ & 988 & 1128 & 1285 & 1465 & 1680 & 1945 & 2229 & 2527 & 2914 & 3130 & 3282 & 3419 & 3563  & 3754 & 4004 & 4096  \\ \hline
    conv$5\_1$ & 1095 & 1265 & 1457 & 1669 & 1840 & 2004 & 2086 & 2143 & 2264 & 2371 & 2485 & 2570 & 2671  & 2813 & 2867 & 2998  \\ \hline
    conv$5\_2$ & 790 & 926 & 1037 & 1160 & 1287 & 1349 & 1413 & 1509 & 1569 & 1634 & 1700 & 1772 & 1830  & 1896 & 1998 & 2074  \\ \hline
    conv$5\_3$ & 457 & 513 & 553 & 594 & 632 & 662 & 693 & 732 & 757 & 793 & 812 & 849 & 893  & 914 & 943 & 987  \\ \hline
  \end{tabular}
    \begin{tablenotes}[para,flushleft]
  \end{tablenotes}
  \end{threeparttable}
\end{table*}

For the cross-layer nonuniform quantization, the exhaustive search is employed. Since there are 13 convolutional layers in the VGG-16, the combination of quantization bits, which achieves the best tradeoff between accuracy and the size of all activations, is (8, 8, 10, 11, 11, 11, 11, 11, 11, 10, 10, 9, 8).
The quantization results on the AlexNet is similar to that in~\cite{lin2015fixed}.



\subsection{ENQ and KNQ Schemes}

For the VGG-16, the numbers of quantization bits generated by the ENQ scheme are shown in Tables~\ref{tab:discrete1} and~\ref{tab:discrete2}, where several combinations of $E_i$'s and the corresponding accuracy are shown. Index 6 is the combination which achieves the best tradeoff between accuracy and the number of required memory bits.

The results generated by the KNQ scheme on VGG-16 and AlexNet are shown in Tables~\ref{tab:centroid} and~\ref{tab: K-means}. As shown in Table~\ref{tab:centroid}, $E_i=5$ for each convolutional layer $i$ in the VGG-16. The corresponding computed 32 centroid fixed-point numbers for each convolutional layer are shown in Table~\ref{tab:centroid}. For the AlexNet, $E_i=3$ for each convolutional layer $i$. The centroid fixed-point numbers are not shown here for simplicity. Besides, for both VGG-16 and AlexNet, the pre-processing can improve the accuracy as shown in Table~\ref{tab: K-means}.

In this paper, the proposed quantization unit and data conversion unit for VGG-16 are implemented based a TSMC 90nm CMOS technology.
\begin{table}[hbt]
  \centering
  \caption{Area and Critical Path Delay}
  \label{tab: hw_implement}
  \footnotesize
  \begin{tabular}{c|c|c|c|c}
    \hline
    & QE&CE&QK&CK \\ \hline\hline
  area ($\mu m^2$)  & 776 & 311 & 7512 &4087   \\ \hline
   CPD (ns)         & 0.43& 0.25 & 0.65& 0.31   \\\hline\hline
  \end{tabular}
\end{table}

\subsection{Comparisons and Discussions} \label{sec: comp}
In Table~\ref{tab: compV}, we compare the proposed ENQ and KNQ schemes with the cross-layer nonuniform quantization scheme~\cite{lin2015fixed} in terms of accuracy and the number of required bits to store all activations. As shown in Table~\ref{tab: compV}, NB denotes the number of bits required to store all activations and NNB denotes the normalized NB. As shown in Table~\ref{tab: compV}, compared to that in~\cite{lin2015fixed}, both of our quantization schemes reduce the number of required memory bits by around 50\% for both considered DCNNs. On the other hand, for the VGG-16, the ENQ and KNQ schemes increase the accuracy by 0.03\% and 0.36\% compared to the scheme in~\cite{lin2015fixed}, respectively. For the AlexNet, the accuracy of ENQ and KNQ schemes is slightly worse than that of the scheme in~\cite{lin2015fixed}. 

\begin{table}[hbt]
  \centering
  \caption{K-means clustering Quantization}
  \label{tab: K-means}
  \footnotesize
  \begin{tabular}{c|c|c|c|c}
    \hline
    &  \multicolumn{2}{c}{no pre-process}   & \multicolumn{2}{|c}{with pre-process} \\ \hline\hline
  Bit-width for VGG-16  & 4 & 5 & 4 & 5   \\ \hline
   Top5 Accuracy of VGG-16($\%$)  & $62.8$   &   $85.8$   &   $69.57$    &    $86.58$     \\\hline\hline
   Bit-width for AlexNet  & 2 & 3 & 2 & 3   \\ \hline
   Top5 Accuracy of AlexNet($\%$)  & $34.41$   &   $77.55$   &   $39.03$    &    $78.23$     \\\hline
  \end{tabular}
\end{table}

\begin{table}[hbt]
  \centering
  \caption{Comparison with ~\cite{lin2015fixed}}
  \label{tab: compV}
  \footnotesize
  \begin{tabular}{c|c|c|c|c}
    \hline
    \multicolumn{5}{c}{VGG-16 Net} \\ \hline\hline
   & Half-precision  & ENQ & KNQ &  ~\cite{lin2015fixed}  \\ \hline
   Accuracy($\%$)   &  $88.4$ &    $86.25$   &   $86.58$  & $86.22$   \\\hline
   NB (MB)    & 27 &  $9.5$   &   $8.4$ &  $16.8$   \\\hline
   NNB  & 1.6 &    $0.56$   &   $0.50$  &  $1$ \\\hline\hline
   \multicolumn{5}{c}{Alex Net} \\ \hline
   & Half-precision  & ENQ & KNQ &  ~\cite{lin2015fixed}  \\ \hline
   Accuracy($\%$)   &  $79.95$ &    $77.55$   &   $78.23$  & $78.87$   \\\hline
   NB (MB)    & 2.57 &  $0.48$   &   $0.48$ &  $1.12$   \\\hline
   NNB   & 2.29 &    $0.43$   &   $0.43$  &  $1$ \\\hline
  \end{tabular}
\end{table}

As shown in Fig.~\ref{fig: vgg-16}, most of activations are 0 after ReLU computing. With various data compression techniques, the number of required memory bits can be reduced in further. Besides, it is believed that the fine-tuning can help improve the accuracy.

\section{Conclusion} \label{sec: conclusion}

In this paper, we have applied several methods on the quantization
of DCNN including the VGG-16 and AlexNet. It has been demonstrated that
only 5 and 3 bits are needed for quantizing activations of the VGG-16
and AlexNet, respectively. Compared to the state-of-the-art quantization
schemes, the proposed ENQ and KNQ schemes achieve significant
reductions on the required memory storage for activations. The memory saving feature of the
proposed schemes will facilitate the embedded
hardware implementations of DCNNs.

\bibliographystyle{IEEEbib}
\bibliography{refs_latest}

\end{document}